\documentclass[]{opendatalab}
\usepackage{xcolor}
\usepackage{longtable}
\usepackage{listings}
\usepackage{array}
\usepackage{makecell}
\usepackage{tabularx}
\usepackage[misc]{ifsym}

\definecolor{codegreen}{rgb}{0,0.6,0}
\definecolor{codegray}{rgb}{0.5,0.5,0.5}
\definecolor{codepurple}{rgb}{0.58,0,0.82}
\definecolor{backcolour}{rgb}{0.95,0.95,0.92}
\lstdefinestyle{mystyle}{backgroundcolor=\color{backcolour},commentstyle=\color{codegreen},keywordstyle=\color{magenta},numberstyle=\tiny\color{codegray},stringstyle=\color{codepurple},basicstyle=\ttfamily\footnotesize,breaklines=true,captionpos=b,keepspaces=true,numbers=none,showspaces=false,showstringspaces=false,showtabs=false,tabsize=2}
\title{MinerU.Chem: A High-Precision System for Optical Chemical Structure and Reaction Recognition}

\author[1\dag]{Haote Yang}
\author[1\dag\ddagger]{Jiang Wu}
\author[2\dag]{Jingchao Wang}
\author[1]{Xingjian Wei}

\author[3]{Lixin Ma}
\author[4]{Linye Li}
\author[5]{Chen Zhu}
\author[3]{Xiaolong Wu}
\author[6]{Yuheng Lu}
\author[1]{Ziran Zhu}
\author[1]{Junyuan Gao}
\author[7]{Lingli Ge}
\author[8]{Yuan Xu}
\author[4]{Huijie Ao}

\author[1]{QianQian Wu}
\author[1]{Dechen Lin}
\author[1]{Huaiyu Gu}

\author[1]{Lu Chen}
\author[1]{Shengxin Lu}
\author[1]{ShaSha Wang}
\author[1]{Yuanyuan Cao}

\author[1]{Zhejia Yu}
\author[1]{Ruijie Zhang}
\author[1]{Zimai Tian}
\author[1]{Jiaxing Sun}

\author[1]{Yinfan Wang}
\author[1,7]{Jiahe Song}
\author[1,7]{Chuang Wang}
\author[1]{Yubin Wang}
\author[9]{Rui Nie}
\author[10]{Hao Zheng}
\author[11]{Bowen Jiang}
\author[11]{Hongbin Lai}

\author[11]{Yifan He}
\author[12]{Chengjin Liu}

\author[1]{Tingting Zhang}
\author[1]{Liqun Wei}

\author[1]{Lijun Wu}
\author[1]{Bin Wang}
\author[1]{Yuqiang Li}
\author[1]{Guangyu Wang}
\author[1]{Wei Li}

\author[1]{Bowen Zhou}
\author[1]{Dahua Lin}
\author[1\ \textrm{\Letter}]{Conghui He}

\affiliation[1]{Shanghai Artificial Intelligence Laboratory}
\affiliation[2]{East China Normal University}
\affiliation[3]{Tongji University}
\affiliation[4]{Fudan University}
\affiliation[5]{East China University of Science and Technology}
\affiliation[6]{Beijing University of Posts and Telecommunications}
\affiliation[7]{Shanghai Jiao Tong University}
\affiliation[8]{Jilin University}
\affiliation[9]{Beihang University}
\affiliation[10]{South China Normal University}
\affiliation[11]{Peking University}
\affiliation[12]{Northwestern Polytechnical University}

\abstract{
In organic chemistry papers and patents, molecular structures, reaction schemes, and experimental conditions are often presented as molecular structure depictions, reaction diagrams, and complex tables or figures. Such information is difficult for general-purpose document parsing systems to directly convert into machine-readable data. This limits data production for organic chemistry knowledge base construction and for AI for Chemistry tasks such as reaction prediction, retrosynthesis, condition recommendation, molecular property prediction, and drug molecule design. This report introduces \textbf{MinerU.Chem}, a document parsing system for organic chemistry literature integrated into the MinerU online platform. Built on top of MinerU's general document parsing pipeline, MinerU.Chem adds five chemistry-specific modules: chemistry relevance filtering, molecular structure detection, molecule identifier extraction, molecular structure recognition, and reaction scheme parsing. Together, these modules convert organic-chemistry-related image regions in documents into a Molecule Summary List and a Reaction Summary List. For molecular structure recognition, MinerU.Chem uses \textbf{CARBON} (Complex Atomic Representation and Bonding Object Notation) as its core representation. CARBON enables recognition results to preserve both the visual layout of the original image and complex chemical semantics, while supporting the export of standard downstream formats such as MolFile and SMILES. On the SMILES-evaluable subset of the latest revised version of MolRecBench-Wild~\citep{yang2026molrecbench} ($N=2{,}392$), MinerU.Chem's molecular structure recognition module achieves a SMILES exact-match accuracy of 93.02\%, substantially outperforming the best comparison system evaluated in this report, GPT-5.6-Sol (74.87\%), by 18.15 percentage points. The system has been integrated into the MinerU online platform and is available at \href{https://mineru.net/OpenSourceTools/Extractor}{MinerU}.
}

\date{\today}
\metadata[Equal Contribution ($\dagger$)]{Haote Yang, Jiang Wu, Jingchao Wang}
\metadata[Project Lead ($\ddagger$)]{Jiang Wu}
\correspondence{Conghui He, \email{heconghui@pjlab.org.cn}}
\metadata[Online Service]{\url{https://mineru.net/OpenSourceTools/Extractor}}

\begin{document}

\maketitle

\section{Introduction}
\label{sec:1-introduction}

\subsection{Challenges and Limitations in Organic Chemistry Image Recognition}
\label{subsec:challenges-and-limitations-in-organic-chemist}

A large amount of critical information in organic chemistry literature is not presented as text, but embedded in molecular structure depictions, reaction schemes, and figure or table regions. Molecular structure depictions encode atom connectivity, visual layout, atom and group labels, stereochemical cues, repeating units, and other information; reaction schemes further organize reactants, products, reaction conditions, and reaction steps. For chemists, these visual elements are the most intuitive carriers of knowledge; for AI for Chemistry, they are also an important source of high-quality structured data.

Existing general-purpose document parsing systems typically treat these regions as ordinary images without parsing their internal chemical semantics. This limitation hinders the construction of organic chemistry knowledge bases and the production of structured data for AI for Chemistry applications. Optical Chemical Structure Recognition (OCSR) and literature-oriented chemical structure extraction systems can recover atom connectivity, i.e., molecular topology, from molecular structure depictions. However, in real-world literature, low-resolution scans, crowded structures, complex backgrounds, non-standard representations, and visually confusable bond types can still significantly reduce recognition reliability~\citep{yang2026molrecbench,qian2023molscribe,chen2024molnextr,rajan2023decimer,xiong2024alphaextractor}. In addition, non-standard chemical semantics such as stereochemistry, repeating units, Markush fragments, and coordination bonds can be simplified or lost when converted into conventional molecular representations~\citep{yang2026molrecbench}. For database construction and model training, a system must recover not only molecular topology but also complex chemical semantics and image coordinates, thereby enabling downstream processing, human verification, and traceability to the source image.

Reaction scheme parsing faces similar challenges. RxnScribe, RxnCaption, RxnIM, and related work have formulated and modeled reaction diagram parsing and reaction image understanding from different perspectives~\citep{qian2023rxnscribe,song2026rxncaption,chen2025rxnim}. However, in real-world literature scenarios involving dense layouts, low-quality scans, complex backgrounds, and cross-region associations, it remains difficult to robustly extract and organize reactants, products, and conditions. Therefore, data production workflows for real organic chemistry literature still require a system that can work together with general document parsing while further understanding the chemical semantics inside molecular structure depictions and reaction schemes.

\subsection{MinerU and MinerU.Chem}
\label{subsec:mineru-and-mineru-chem}

MinerU is a general-purpose parsing tool for scientific documents that converts PDFs and images into structured Markdown and JSON for LLM, RAG, and agent workflows. Its pipeline performs layout analysis, text recognition, formula parsing, and table parsing, producing structured outputs linked to their locations in the source document~\citep{wang2024mineru,niu2025mineru25,wang2026mineru25pro}. Users can upload documents to the MinerU online platform and download the resulting parsed content for further processing.

Building on the document-level Markdown and layout outputs produced by MinerU, MinerU.Chem applies chemistry-aware parsing to relevant image regions in organic chemistry documents. It generates two structured outputs linked to their source locations:

\begin{enumerate}
\item \textbf{Molecule Summary List}: records molecules detected and recognized in the document, together with their source locations, identifiers, MolFile, and SMILES.
\item \textbf{Reaction Summary List}: organizes parsed reaction diagrams into reactants, products, and conditions, with links to corresponding molecule records whenever possible.
\end{enumerate}

The chemistry-aware layer comprises five modules: chemistry relevance filtering, molecular structure detection, molecule identifier extraction, molecular structure recognition, and reaction scheme parsing. Together, these modules transform general document-parsing outputs into verifiable and traceable structured chemical information for downstream data production.

\subsection{Faithful Representation of Molecular Structures}
\label{subsec:faithful-representation-of-molecular-structur}

Among the chemistry-specific capabilities of MinerU.Chem, molecular structure recognition is central to downstream data quality. The module must produce representations compatible with cheminformatics tools while faithfully preserving the complex structural semantics encoded in the source image.

SMILES~\citep{weininger1988smiles} is the most widely used molecular representation in cheminformatics and is suitable for expressing atom connectivity. However, it has limited ability to represent non-standard bonds, atom attributes (such as valence states, radicals, and attachment points), repeating units, Markush fragments, and the visual layout of the original image. Alternative and extended molecular representations, including SELFIES~\citep{krenn2020selfies}, CSMILES~\citep{furness2025csmiles}, MolFile, and E-SMILES~\citep{fang2025molparser}, address different requirements, such as syntactic robustness, interoperability, and richer structural semantics. Faithful molecular structure recognition in real-world literature therefore requires recovering not only conventional SMILES-compatible connectivity, but also structural semantics and depiction-specific information that conventional SMILES does not fully preserve~\citep{yang2026molrecbench}.

MinerU.Chem adopts \textbf{CARBON} (Complex Atomic Representation and Bonding Object Notation) as the core data representation in its molecular structure recognition module. CARBON was introduced by MolRecBench-Wild~\citep{yang2026molrecbench}. It is an atom-centric graph representation for molecular structure depictions that can organize chemical semantic information such as atoms, bonds, abbreviations, non-standard bonds, and repeating structures, while preserving the two-dimensional coordinates of atoms in the original image~\citep{yang2026molrecbench}. MinerU.Chem extends CARBON from an evaluation representation into the native prediction format of its deployed molecular structure recognition module. The system preserves CARBON internally as the primary representation and derives standard formats, including MolFile and SMILES, for compatibility with downstream cheminformatics toolchains. This design aligns recognition results with the visual layout of the source image, preserves complex chemical semantics, and maintains interoperability with established molecular data formats.

\subsection{Main Contributions}
\label{subsec:main-contributions}

The main contributions of this report are as follows:

\begin{enumerate}
\item \textbf{Integrated deployment of chemistry parsing capabilities.} MinerU.Chem integrates chemistry relevance filtering, molecular detection, identifier extraction, structure recognition, and reaction scheme parsing into the MinerU~\citep{wang2024mineru,niu2025mineru25,wang2026mineru25pro} online platform, generating Molecule Summary Lists and Reaction Summary Lists associated with locations in the original document.
\item \textbf{Systematic deployment of CARBON.} MinerU.Chem adopts CARBON, the target representation in the MolRecBench-Wild~\citep{yang2026molrecbench} evaluation protocol, as the native prediction format of the molecular structure recognition module. CARBON preserves molecular layout and chemical semantics that conventional SMILES cannot fully cover, while the online system exports recognized structures in standard MolFile and SMILES formats for downstream use.
\item \textbf{High-performance molecular recognition for real-world literature complexity.} On the latest revised version of MolRecBench-Wild~\citep{yang2026molrecbench}, MinerU.Chem's molecular structure recognition module achieves a SMILES exact-match accuracy of 93.02\% on the SMILES-evaluable subset ($N=2{,}392$) and a graph accuracy of 79.66\% on all graph-annotated samples ($N=5{,}024$). These results outperform the best comparison systems evaluated in this report by 18.15 percentage points in SMILES exact-match accuracy (GPT-5.6-Sol, 74.87\%) and 43.25 percentage points in graph accuracy (Gemini-3.5-flash-thinking, 36.41\%).
\end{enumerate}

\begin{figure}[tbp]
\centering
\includegraphics[width=0.97\linewidth]{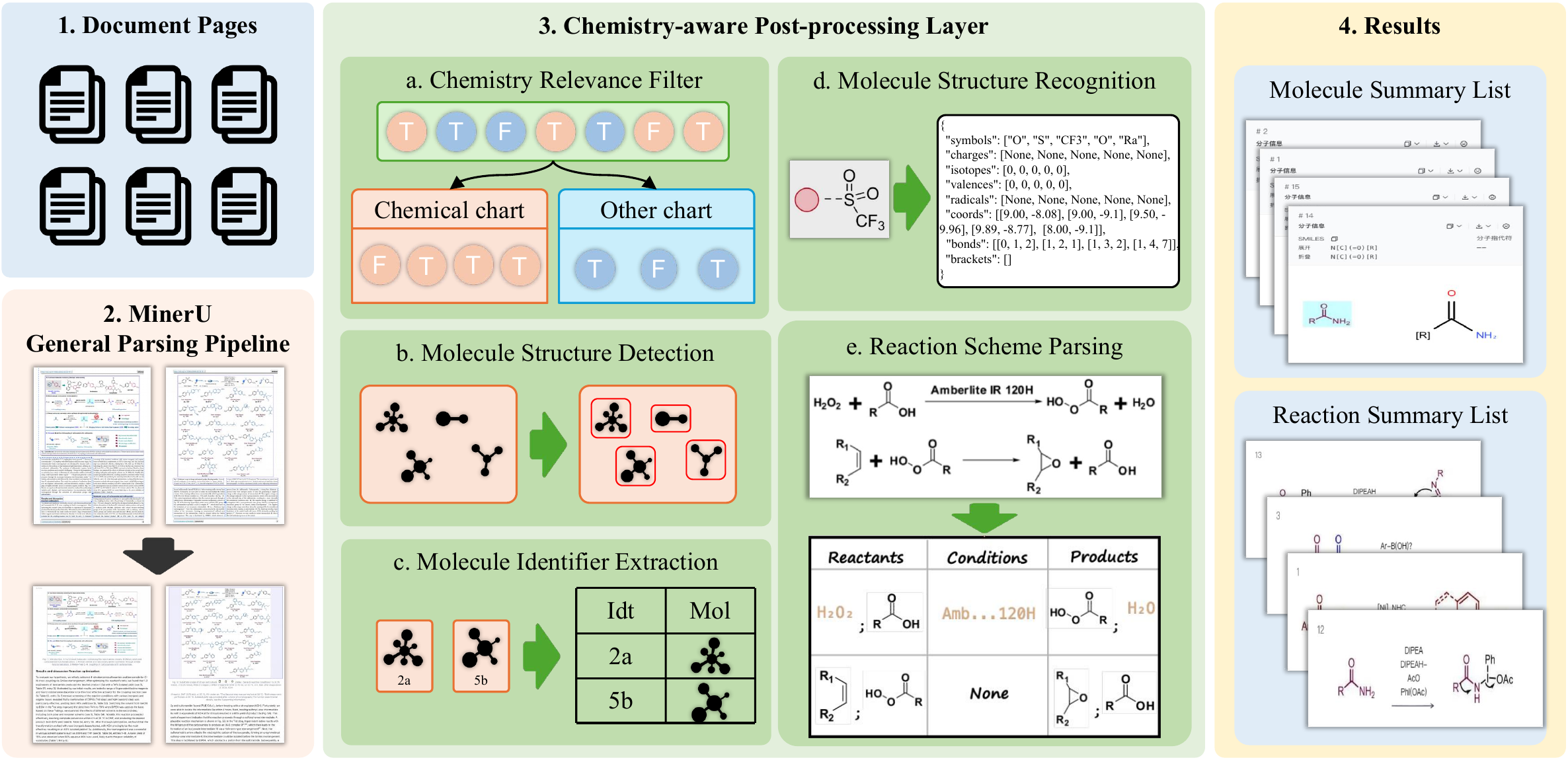}
\caption{The overall workflow of MinerU.Chem.}
\label{fig:mineru-chem}
\end{figure}

\section{System Overview}
\label{sec:2-system-overview}

\subsection{Overall Workflow}
\label{subsec:overall-workflow}

MinerU.Chem operates as a chemistry-aware post-processing layer within the MinerU pipeline. MinerU first converts PDFs or page images into document-level Markdown and layout outputs; MinerU.Chem then parses molecule- and reaction-related regions to produce structured chemical records. Figure~\ref{fig:mineru-chem} illustrates the end-to-end workflow.

\begin{figure}[tbp]
\centering
\includegraphics[width=0.8\linewidth]{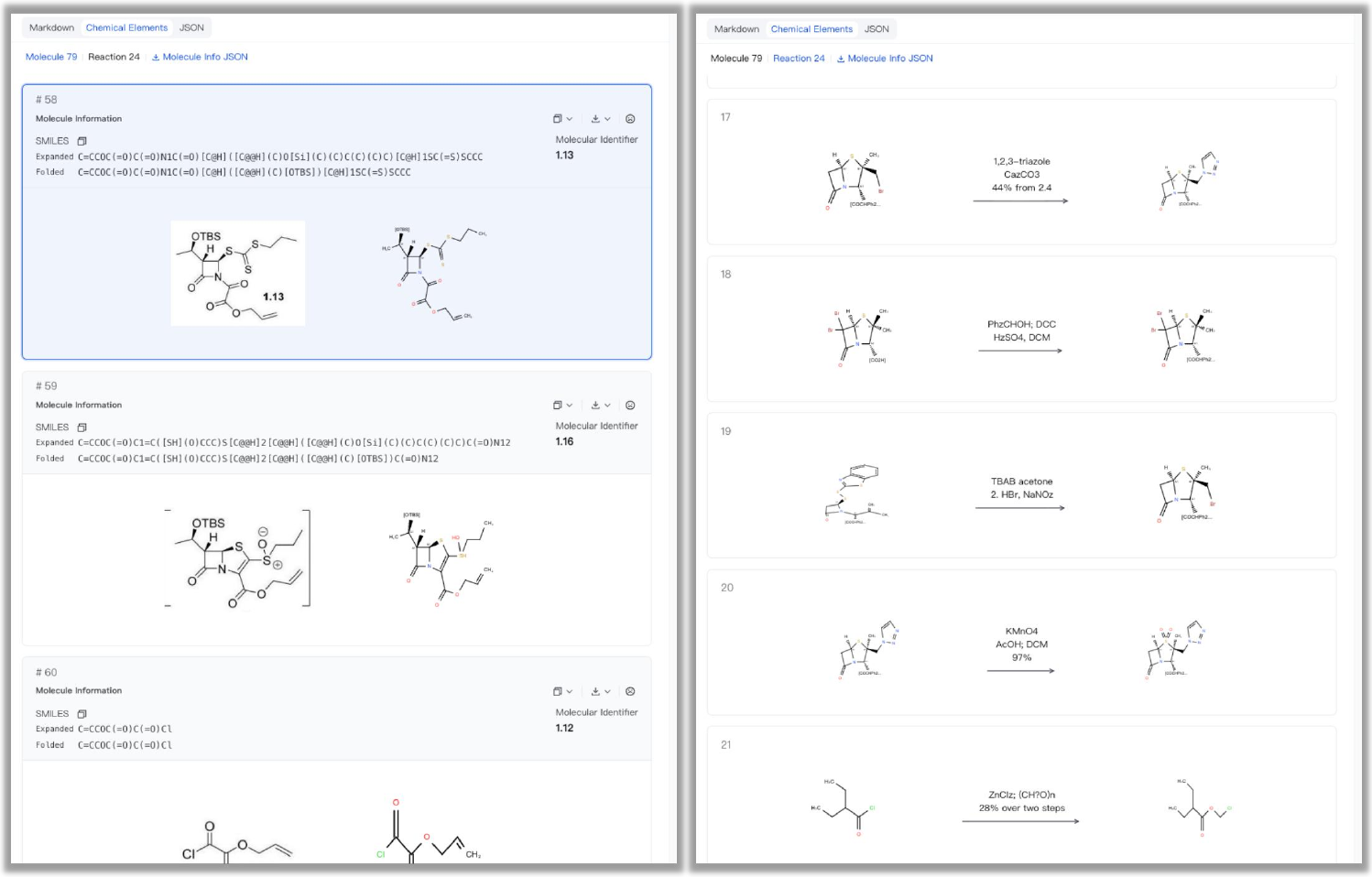}
\caption{An example of the Molecule Summary List (left) and the Reaction Summary List (right).}
\label{fig:two-types-output}
\end{figure}

MinerU.Chem produces two complementary structured outputs: the Molecule Summary List and the Reaction Summary List (Figure~\ref{fig:two-types-output}). Both are linked to evidence in the source document through page numbers, within-page bounding boxes, identifiers, and cross-references between molecules and reactions. The Molecule Summary List consolidates the molecules detected and recognized throughout the document, recording the source location, identifier, MolFile, and SMILES for each molecule. Building on these molecule records, the Reaction Summary List organizes parsed reaction diagrams into reactants, products, and conditions. Whenever possible, reactants and products are linked to their corresponding molecule records, enabling each reaction component to be traced back to its recognized structure and source location.

\subsection{Chemistry Parsing Modules}
\label{subsec:chemistry-parsing-modules}

MinerU.Chem consists of five chemistry-specific parsing modules, summarized in Table~\ref{tab:mineru-chem-module}. Together, these modules generate consistently structured Molecule Summary Lists and Reaction Summary Lists within the document context. The remaining four modules---molecular structure detection, molecule identifier extraction, molecular structure recognition, and reaction scheme parsing---are described in Sections~\ref{sec:3-molecular-structure-detection}--\ref{sec:6-reaction-scheme-parsing}.

\begin{table}[t]
\centering
\caption{The parsing modules of MinerU.Chem.}
\label{tab:mineru-chem-module}
\small
\begin{tabular}{p{0.28\textwidth}p{0.64\textwidth}}
\toprule
\textbf{Module} & \textbf{Function} \\
\midrule
\textbf{Chemistry Relevance Filter} & Performs binary classification on figure/table regions extracted from a page, determines whether each region is chemistry-related, and routes only positive regions to downstream modules. \\
\hline
\textbf{Molecular Structure Detection} & Locates molecular structure regions and outputs molecule bounding boxes. \\
\hline
\textbf{Molecule Identifier Extraction} & Associates molecular structure depictions with nearby compound numbers, labels, or other identifiers. \\
\hline
\textbf{Molecular Structure Recognition} & Converts molecular structure depictions into CARBON graph representations and derives standard formats such as MolFile and SMILES. \\
\hline
\textbf{Reaction Scheme Parsing} & Parses reaction schemes into reaction entries containing reactants, products, and conditions. \\
\bottomrule
\end{tabular}
\end{table}

\subsection{User Experience Overview}
\label{subsec:user-experience-overview}

MinerU.Chem is integrated into the MinerU online platform as an optional chemistry-aware parsing mode. Users can activate it by enabling the "Chemistry Paper" option in the settings at the upper-right corner of the parsing page. Once activated, MinerU.Chem generates structured chemical outputs in addition to the standard Markdown and layout-parsing results.

Users can therefore directly inspect recognition results, trace them back to evidence in the original document, and download summary tables for human verification, database construction, and downstream AI for Chemistry data production.

\section{Molecular Structure Detection}
\label{sec:3-molecular-structure-detection}

Molecular structure detection locates molecular structures on document pages and within figure or table regions and outputs their corresponding within-page bounding boxes. The module determines where molecular structures appear but does not identify the molecules, recognize their chemical structures, or assign their roles in reactions. Its outputs serve as inputs for downstream molecule identifier extraction, molecular structure recognition, and reaction scheme parsing.

Molecule localization in real-world documents presents several challenges. First, molecular structures vary substantially in scale and layout, ranging from small isolated molecules to large structures occupying much of a page and densely arranged structures in reaction schemes or multi-molecule panels. Second, low-resolution scans, blur, and poor contrast can obscure bond lines, atom labels, and structural boundaries. Third, arrows, condition labels, legends, table content, and other surrounding elements may appear adjacent to or overlap with molecular structures, making their boundaries difficult to determine. Missed detections, false positives, or bounding-box shifts at the detection stage can further affect downstream identifier pairing, structure recognition, and reaction organization~\citep{rajan2021decimerseg,zhou2024yode}.

MinerU.Chem uses MolYOLO, introduced in the team's prior work RxnCaption~\citep{song2026rxncaption}, to detect candidate molecular structure regions on document pages and within figure or table regions. Table~\ref{tab:moldet_performance} reports the MolDet-33k-test results originally presented in RxnCaption.

\begin{table}[t]
    \centering
    \caption{Performance comparison of molecular structure detection models on MolDet-33k-test.}
    \label{tab:moldet_performance}
    \begin{tabular}{lcc}
        \toprule
        Model & Precision@IoU 0.5 & Recall@IoU 0.5 \\
        \midrule
        MolDetect      & 0.84 & 0.77 \\
        YoDe           & 0.89 & 0.75 \\
        MolYOLO (ours) & \textbf{0.98} & \textbf{0.98} \\
        \bottomrule
    \end{tabular}
\end{table}

\section{Molecule Identifier Extraction}
\label{sec:4-molecule-identifier-extraction}

Molecule identifier extraction aims to associate detected molecular structure depictions with nearby compound numbers or labels. Here, identifiers typically refer to compound numbers or labels placed next to molecular structures, such as \texttt{1}, \texttt{2}, \texttt{3}, \texttt{1a}, and \texttt{2a}. Identifiers themselves are not final chemical data, but they are important anchors for document-level molecule organization: through these anchors, the system can track and reference specific molecular structures across reaction schemes, figure/table regions, and full-document contexts.

\begin{figure}[tbp]
\centering
\includegraphics[width=\linewidth]{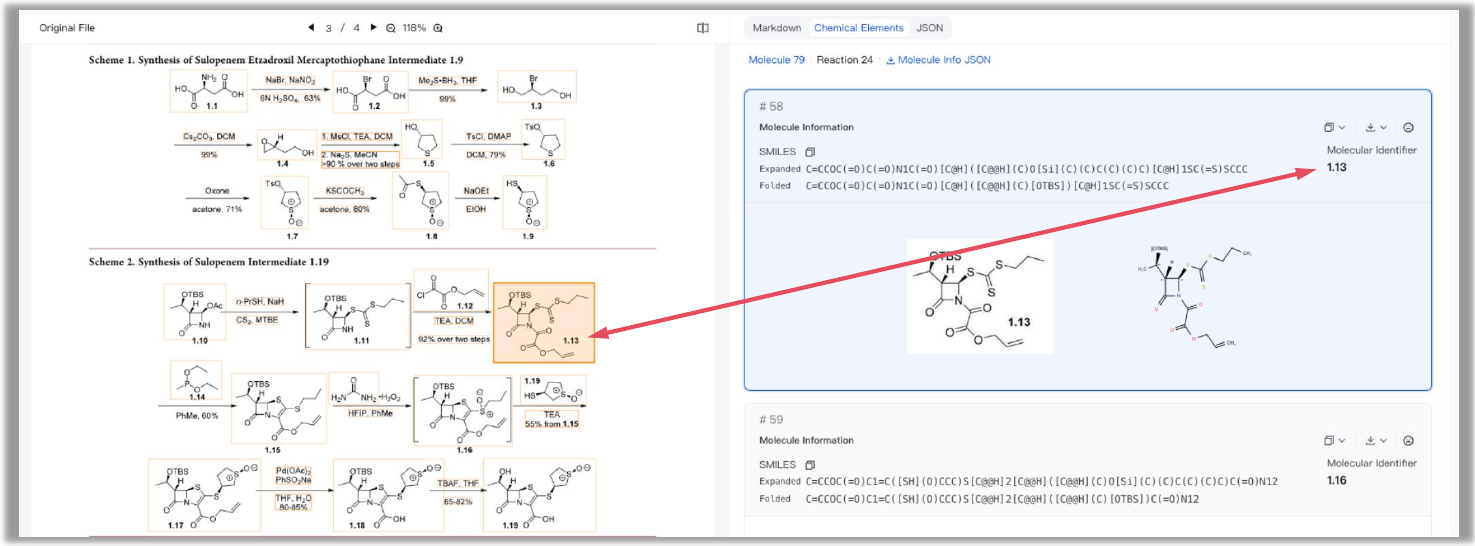}
\caption{An example of the molecule-identifier pairing.}
\label{fig:idt}
\end{figure}

Based on detected molecular regions and nearby text or annotations, this module outputs molecule-identifier pairings. The pairings are written into the Molecule Summary List, so that the list is not merely an isolated structure list, but can be linked to reaction entries and document references. For example, when reactants or products in a reaction scheme appear as numbered compounds, the system can use identifiers to trace them back to the corresponding molecular structure records. Figure~\ref{fig:idt} shows an example of the molecule-identifier pairing.

This module builds on the Mid-Mapper (Molecule-Identifier Mapper) introduced in the team's prior work, RxnID~\citep{rxnid2026}, while adapting the identifier definition to the requirements of the deployed system. Rather than attempting to cover all complex naming conventions, the current version prioritizes frequently occurring and reliably identifiable compound numbers and labels in scientific papers, thereby improving the robustness of document-level molecule organization.

\section{Molecular Structure Recognition}
\label{sec:5-molecular-structure-recognition}

Molecular structure recognition is the core chemistry capability of MinerU.Chem. It converts detected molecular structure depictions into machine-readable structural representations and writes them into the Molecule Summary List.

\subsection{Recognition Objective: Faithful Structure Reconstruction}
\label{subsec:recognition-objective-faithful-structure-reco}

Rather than reducing each molecular image to a single SMILES string, the molecular structure recognition module aims to recover the structural information encoded in the source image, thereby supporting human verification, database curation, and dataset construction for AI for Chemistry.

Faithful structure reconstruction is reflected in three aspects:

\begin{enumerate}
\item \textbf{Visual fidelity.} Predictions preserve the spatial layout of the source molecular depiction and retain image-aligned coordinates, facilitating inspection and correction of atom- and bond-level outputs.
\item \textbf{Chemical-semantic fidelity.} The representation captures semantics that conventional SMILES cannot fully express, including non-standard bonds, abbreviated structures, repeating units, Markush fragments, and stereochemistry, thereby reducing information loss during structure reconstruction.
\item \textbf{Robustness to real-world variation.} Training and evaluation cover both visual and chemical-semantic challenges found in real-world literature, rather than focusing only on clean, isolated molecular depictions with regular layouts.
\end{enumerate}

\subsection{Representation Design: Graph-level Output Based on CARBON}
\label{subsec:representation-design-graph-level-output-base}

MinerU.Chem follows an image-to-graph recognition approach. Rather than generating SMILES or other string representations directly from an image, the model predicts atoms, bonds, and image-aligned spatial information as a molecular graph. This graph-level representation captures molecular topology, atom- and bond-level attributes, and the spatial layout of the source depiction.

\begin{figure}[tbp]
\centering
\includegraphics[width=\linewidth]{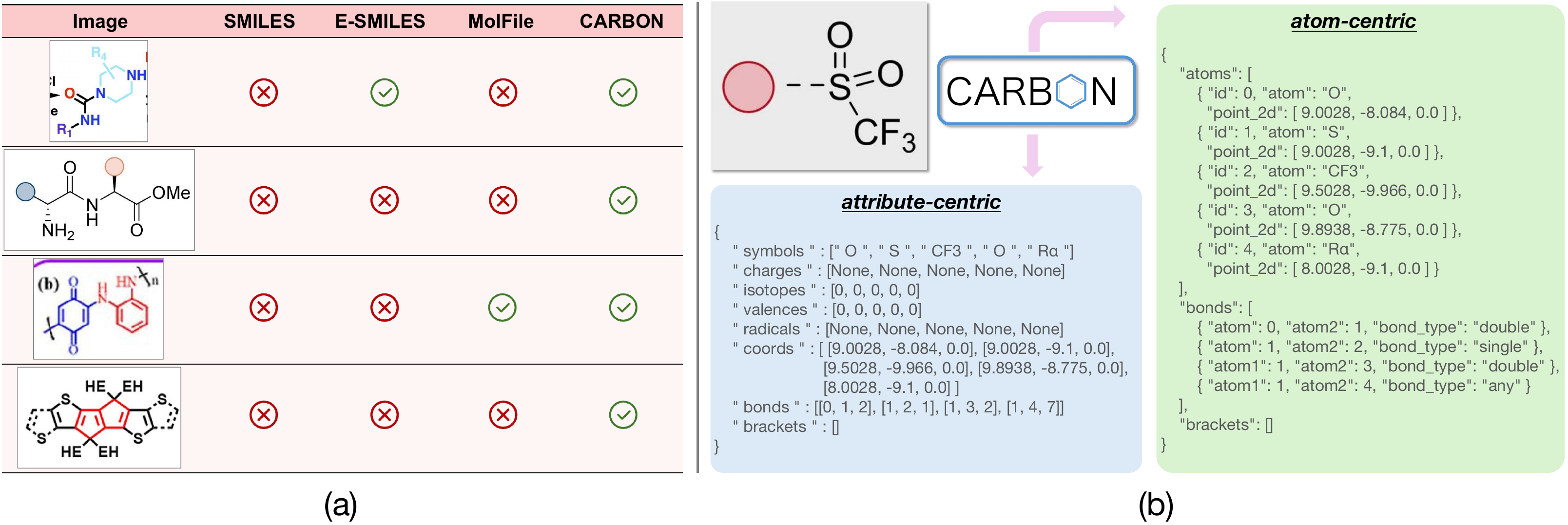}
\caption{Comparison of molecular representation methods: (a) representative molecular representations and (b) an example of the CARBON format.}
\label{fig:carbon}
\end{figure}

Internally, the system uses CARBON as its core representation. CARBON is an atom-centric graph representation for molecular structure depictions that can organize chemical semantic information such as atoms, bonds, abbreviations, non-standard bonds, and repeating structures, while preserving atom coordinates in the original image~\citep{yang2026molrecbench}. In CARBON, abbreviated groups in molecular structure depictions are modeled as superatoms to preserve both the visual layout and structural semantics of the original image. As shown in Figure~\ref{fig:carbon}, compared with SMILES, E-SMILES, and MolFile~\citep{weininger1988smiles,fang2025molparser}, CARBON provides greater flexibility for representing complex chemical semantics and image-aligned spatial information.

\begin{table}[tbp]
\centering
\caption{Output forms of representative OCSR systems.}
\label{tab:output-form}
\small
\begin{tabular}{lccc}
\toprule
System & Output format & \makecell{Graph \\ representation} & SMILES-first \\
\midrule
DECIMER~\citep{rajan2023decimer} & SMILES caption & No & Yes \\
Uni-Parser MolParser 1.5~\citep{dptech2025uniparser} & E-SMILES caption & No & Yes \\
MolScribe~\citep{qian2023molscribe} & SMILES + atom-level annotation & Partial & Yes \\
MolMole / ViMore~\citep{lgai2025molmole} & Detection-based MolFile & Yes & Partial \\
MinerU.Chem & \makecell[l]{Native CARBON graph; \\ exported MolFile / SMILES} & Internal & No \\
\bottomrule
\end{tabular}
\end{table}

In practical applications, downstream databases and cheminformatics toolchains typically require complete atom-level molecular representations. MinerU.Chem therefore uses a post-processing mechanism that expands CARBON superatoms into their corresponding atom-level structures. The recognition module retains the original CARBON representation internally to preserve visual layout and complex structural semantics, while exporting derived MolFile and SMILES formats for downstream interoperability. Table~\ref{tab:output-form} compares the native representations and exported formats of representative OCSR systems.

\subsection{Training Data Composition}
\label{subsec:training-data-composition}

The performance of molecular structure recognition depends not only on representation design, but also on how well the training data covers the complexity of real-world literature. The training data for MinerU.Chem consists of three components: curated open-source data, human-annotated data from real-world literature, and synthetic data targeting challenging scenarios. These components provide broad structural coverage, supervision from real-world literature, and systematic coverage of long-tail difficult cases, respectively.

\begin{figure}[tbp]
\centering
\includegraphics[width=\linewidth]{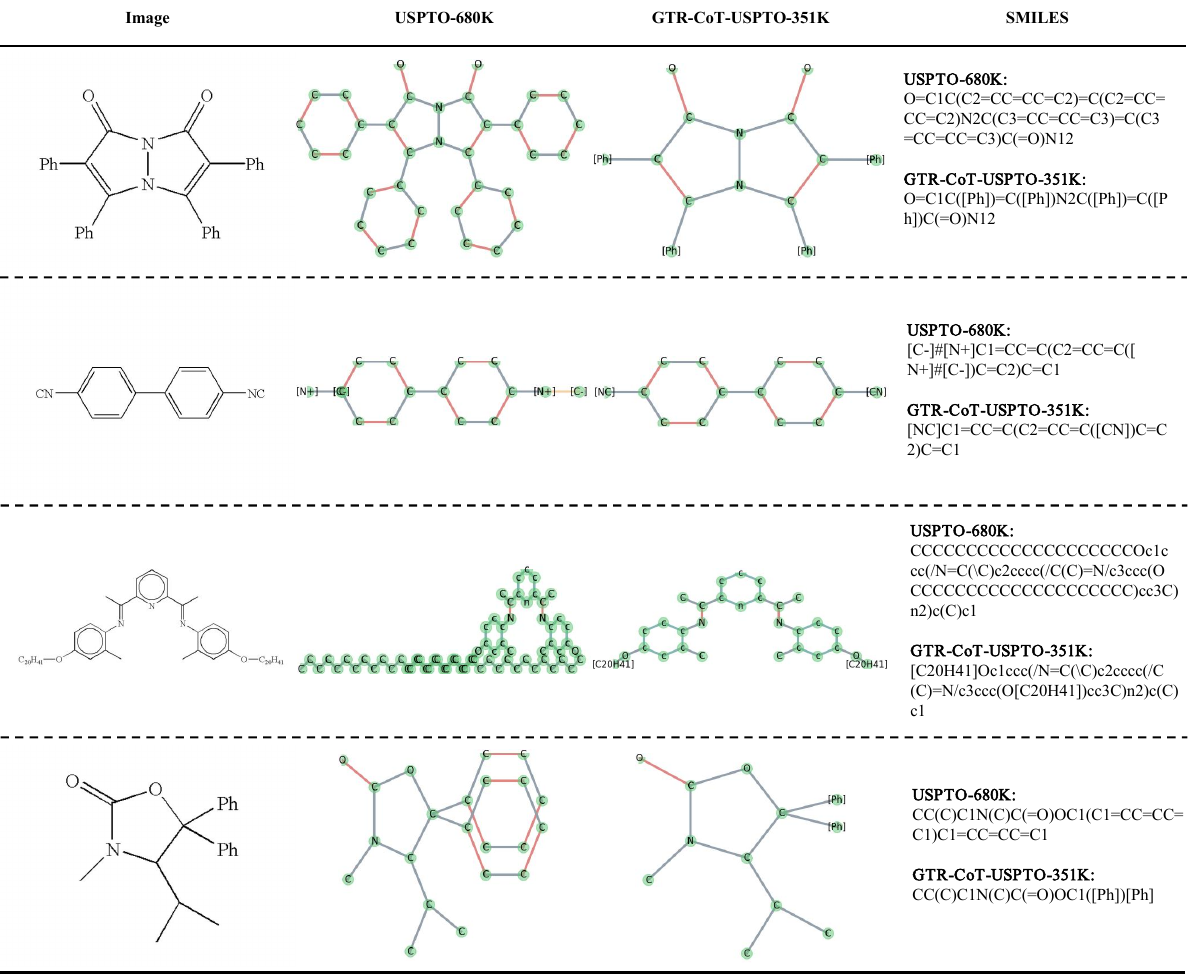}
\caption{Cleaning, normalization, and CARBON representation conversion workflow for open-source molecular structure recognition data.}
\label{fig:open-ocsr-dataset}
\end{figure}

The first component is curated open-source data derived from existing molecular structure recognition datasets. Through cleaning, normalization, and representation conversion, these data are adapted to the CARBON graph representation and the current model-training requirements. This component provides broad coverage of molecular structures and supports the model's foundational recognition capabilities. Figure~\ref{fig:open-ocsr-dataset} shows examples of open-source data used for training.

\begin{figure}[!tbp]
\centering
\includegraphics[width=0.95\linewidth]{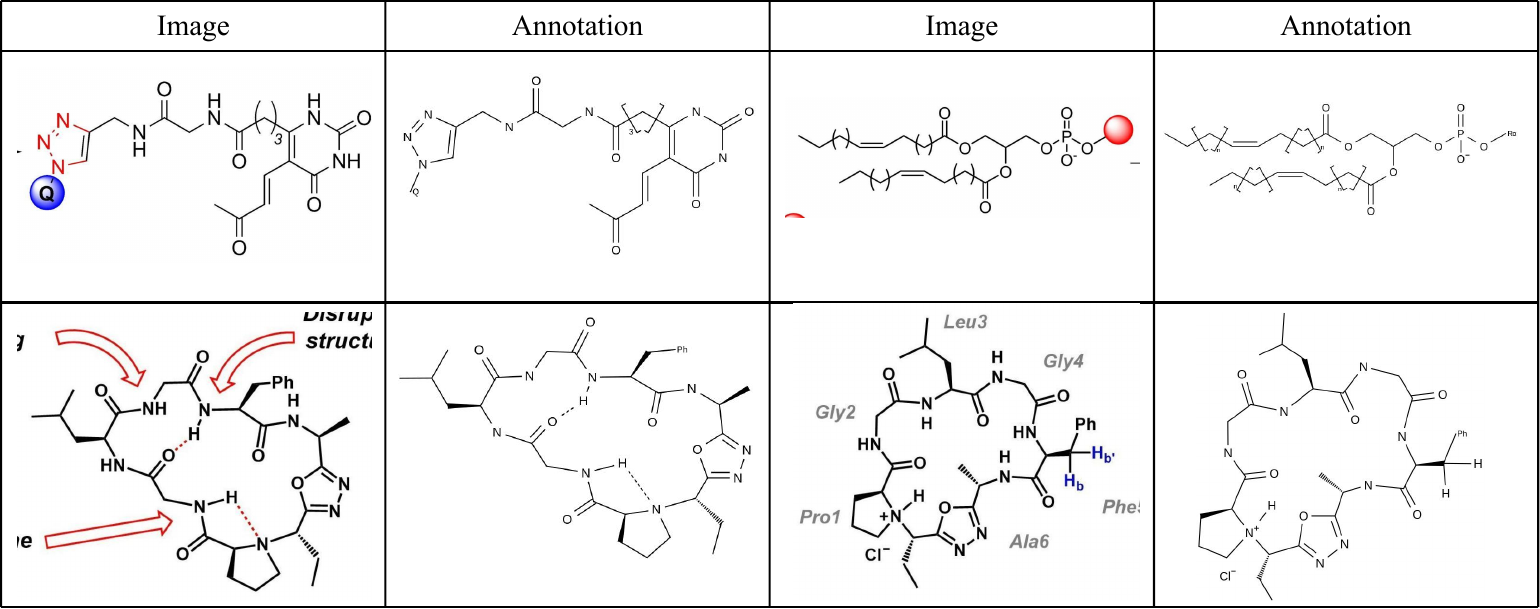}
\caption{Example of fine-grained human annotation of molecular structure depictions from real-world literature.}
\label{fig:ann-ocsr-dataset}
\end{figure}

The second component comprises molecular structure depictions manually annotated from real-world organic chemistry literature using the annotation protocol of MolRecBench-Wild~\citep{yang2026molrecbench}. These annotations provide supervision for atoms, bonds, coordinates, and complex structural semantics, improving the model's reliability on real-world document images, structurally complex molecules, and long-tail chemical representations. Figure~\ref{fig:ann-ocsr-dataset} shows examples of human-annotated data used for training.

\begin{figure}[!tp]
\centering
\includegraphics[width=0.95\linewidth]{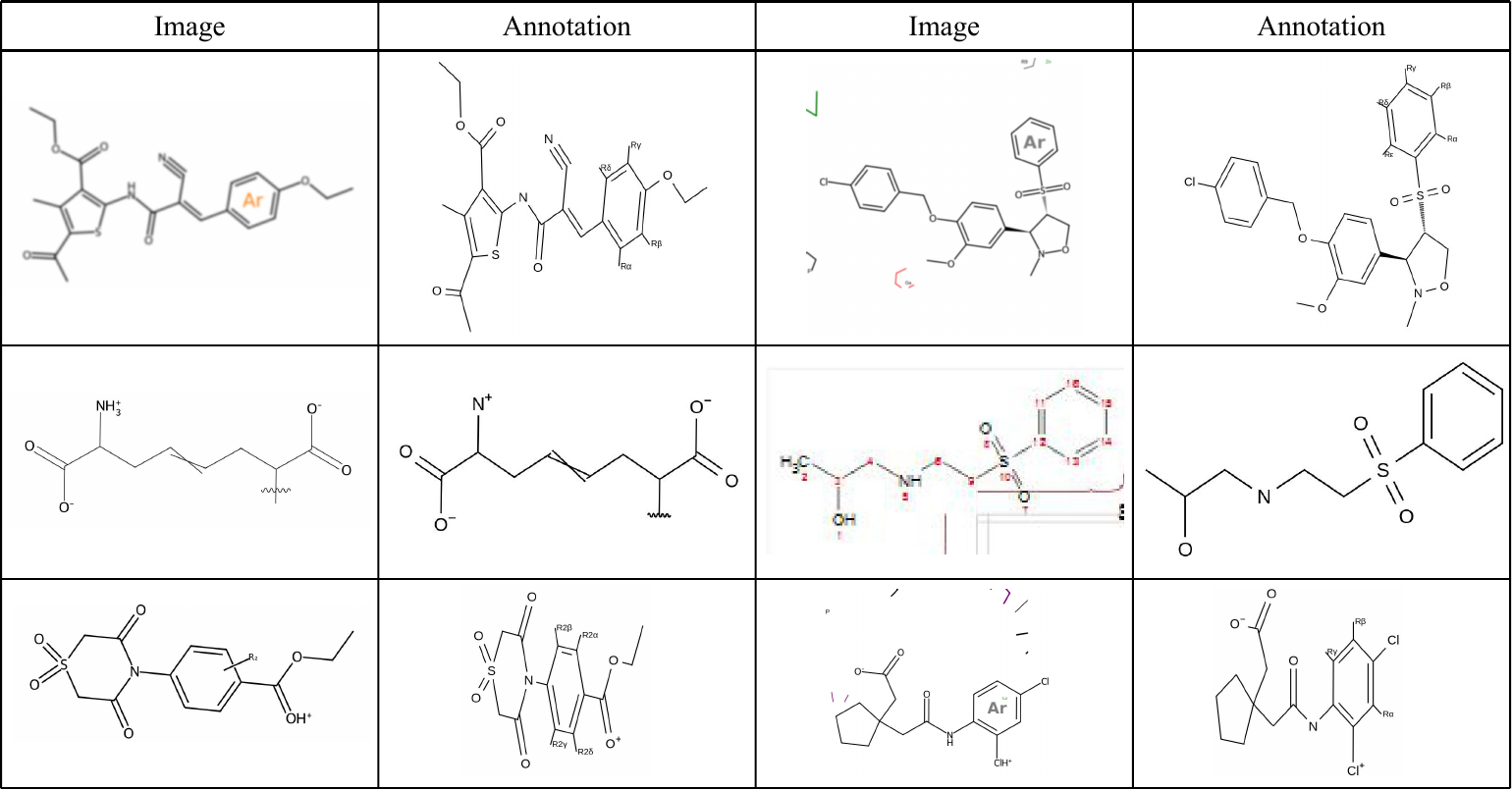}
\caption{Examples of synthetic training data generated according to the two-dimensional MOSAIC difficulty framework.}
\label{fig:syn-ocsr-dataset}
\end{figure}

The third component comprises synthetic data targeting challenging scenarios. A data synthesis engine is used to systematically cover common sources of recognition failure in real-world literature. Following the two-dimensional \textbf{MOSAIC difficulty framework} introduced by MolRecBench-Wild~\citep{yang2026molrecbench}, the synthesis process covers 18 categories of visual complexity---including dense reaction diagrams, crowded table regions, low-quality scans, and decorative elements---and 19 categories of chemical complexity---including stereochemistry, Markush structures, polymers and repeating units, non-standard bonds, abbreviated structures, and coordination bonds---for a total of 37 difficulty labels. This component broadens the coverage of long-tail challenges in the training distribution, enabling the model to handle both clean, regularly laid-out molecular depictions and structurally or visually complex cases from real-world literature. Figure~\ref{fig:syn-ocsr-dataset} shows examples of synthetic data used for training.

Overall, curated open-source data provides broad structural coverage, human-annotated data provides supervision from real-world literature, and synthetic data systematically covers long-tail visual and chemical challenges defined by MOSAIC. Together, these complementary data sources support the molecular structure recognition module's ability to faithfully reconstruct structures from real-world chemical documents.

\subsection{Evaluation Method and Results}
\label{subsec:evaluation-method-and-results}

The molecular structure recognition module of MinerU.Chem builds on the approach described in the previously released GTR-VL technical report~\citep{wang2026gtrvl}. The model has since been further developed using expanded training data, fine-grained annotations from real-world literature, and synthetic data targeting challenging scenarios, resulting in GTR-VL-1.4.13, the version currently deployed online. This report evaluates the online model on MolRecBench-Wild~\citep{yang2026molrecbench} following its multi-track evaluation protocol and reports SMILES exact-match accuracy and graph accuracy. The SMILES metric enables direct comparison with conventional OCSR systems, whereas graph accuracy evaluates exact agreement in molecular topology and atom- and bond-level attributes. Graph accuracy does not directly evaluate two-dimensional coordinates and therefore complements, rather than measures, the image-aligned spatial information preserved by CARBON.

We use the latest revised version of MolRecBench-Wild~\citep{yang2026molrecbench}, which contains 5,024 molecular structure images, all with graph annotations. Among them, 2,392 samples also have valid SMILES annotations and constitute the SMILES-evaluable subset. Accordingly, SMILES exact-match accuracy is computed on these 2,392 samples, whereas graph accuracy is computed on all 5,024 samples. The SMILES-evaluable subset is smaller because some molecular structures contain Markush structures, unconventional abbreviations, illegal direction keys, or bonds that violate standard valence constraints and therefore cannot be converted into valid SMILES.

To further analyze model performance under different sources of difficulty, we divide the benchmark into three mutually exclusive subsets based on its MOSAIC difficulty annotations.
MOSAIC describes molecular structure recognition difficulties along two dimensions: visual presentation and chemical semantics. Based on the MOSAIC difficulty annotations, the benchmark is partitioned into three subsets with progressively greater levels of visual and chemical-semantic complexity:

\begin{itemize}
\item \textbf{Subset A}: samples that do not contain specified chemical semantic property difficulty labels and have relatively few visual difficulty labels. It contains 1,987 images, 1,987 graph annotations, and 1,219 SMILES annotations.

\item \textbf{Subset B}: samples that do not contain the above chemical semantic property difficulty labels but have more visual difficulty labels. It contains 1,976 images, 1,976 graph annotations, and 875 SMILES annotations.

\item \textbf{Subset C}: samples that contain specified chemical semantic property difficulty labels. It contains 1,061 images, 1,061 graph annotations, and 298 SMILES annotations.
\end{itemize}

Subsets A, B, and C contain 1,987, 1,976, and 1,061 molecules, respectively, and together constitute the full MolRecBench-Wild benchmark.

The comparison systems cover three categories: specialized OCSR systems and VLMs fine-tuned for chemistry tasks (OCSU~\cite{fan2025ocsu}, DECIMER v2.2~\cite{rajan2020decimer}, MolGrapher~\cite{morin2023molgrapher}, MolNexTR~\cite{chen2024molnextr}, MolScribe~\cite{qian2023molscribe}, ChemDFM-X~\cite{Zhao2024ChemDFMX}, and ChemVLM~\cite{li2025chemvlm}), commercial chemical OCR services (Logic-Parsing~\cite{chen2025logics}, Mathpix\footnote{\url{https://mathpix.com/}}), and general multimodal large models (InternVL3.5~\cite{internvl2025internvl35}, GLM-4.5V~\cite{glm2025glm45v}, Intern-S1~\cite{bai2025intern}, Seed1.6-Thinking\footnote{\url{https://seed.bytedance.com/zh/blog/introduction-to-techniques-used-in-seed1-6}}, GPT-5.6-Sol\footnote{\url{https://openai.com/index/gpt-5-6/}}, Claude-opus-4-8\footnote{\url{https://www.anthropic.com/news/claude-opus-4-8}}, Gemini-3.5-flash-thinking\footnote{\url{https://deepmind.google/models/model-cards/gemini-3-5-flash}}.

\begin{table}[tbp]
\centering
\caption{Evaluation results of representative methods and systems on MolRecBench-Wild. For the Full, A, B, and C columns, SMILES exact-match accuracy is computed on $N=2{,}392$, $1{,}219$, $875$, and $298$ samples, respectively, and graph accuracy is computed on $N=5{,}024$, $1{,}987$, $1{,}976$, and $1{,}061$ samples, respectively.}
\label{tab:ocsr-metric}
\small
\begin{tabular}{l|cc|cc|cc|cc}
\Xhline{1.5pt}
\multirow{2}{*}{Method}                          & \multicolumn{2}{c|}{Full}        & \multicolumn{2}{c|}{A}           & \multicolumn{2}{c|}{B}           & \multicolumn{2}{c}{C}           \\
\cline{2-9}
                    & SMILES   & Graph          & SMILES   & Graph          & SMILES    & Graph          & SMILES    & Graph          \\
    \midrule
    OCSU                         & 11.41 & -- & 14.60 & -- &  9.03 & -- &  5.37 & -- \\
    DECIMER v2.2                 & 41.43 & -- & 60.21 & -- & 22.97 & -- & 18.79 & -- \\
    MolGrapher                   & 34.78 & -- & 47.01 & -- & 27.77 & -- &  5.37 & -- \\
    MolNexTR                     & 62.50 & -- & 76.78 & -- & 52.11 & -- & 34.56 & -- \\
    MolScribe                    & 62.29 & -- & 77.28 & -- & 50.97 & -- & 34.23 & -- \\
    ChemDFM-X                    & 19.06 & -- & 25.18 & -- & 13.94 & -- &  9.06 & -- \\
    ChemVLM                      &  8.03 & -- & 11.07 & -- &  5.94 & -- &  1.68 & -- \\
    \midrule
    Logic-Parsing                & 25.84 & -- & 33.39 & -- & 21.49 & -- &  7.72 & -- \\
    Mathpix                      & 47.32 & -- & 58.65 & -- & 41.03 & -- & 19.46 & -- \\
    \midrule
    InternVL3.5                  & 39.80 &  3.01 & 46.92 &  4.73 & 34.86 &  2.13 & 25.17 &  1.41 \\
    GLM-4.5V                     & 20.28 &  4.20 & 24.53 &  7.15 & 18.17 &  2.94 &  9.06 &  1.04 \\
    Intern-S1                    & 30.02 &  3.46 & 36.10 &  5.89 & 25.26 &  2.13 & 19.13 &  1.41 \\
    Seed1.6-Thinking             & 24.83 &  4.62 & 30.19 &  7.15 & 19.77 &  3.44 & 17.79 &  2.07 \\
    Claude-opus-4-8              & 65.47 & 14.29 & 72.85 & 21.24 & 61.49 & 11.44 & 46.98 &  6.60 \\
    Gemini-3.5-flash-thinking    & 66.85 & \underline{37.56}
                                             & 72.03 & \underline{50.73}
                                             & 62.97 & \underline{31.93}
                                             & 57.05 & \underline{23.37} \\
    GPT-5.6-Sol                  & \underline{74.87} & 32.56
                                  & \underline{81.95} & 44.44
                                  & \underline{70.06} & 28.14
                                  & \underline{60.07} & 18.57 \\
    \midrule
    \makecell[l]{\textbf{MinerU.Chem}\\\textbf{(GTR-VL-1.4.13)}}
                                 & \textbf{93.02} & \textbf{79.66}
                                 & \textbf{98.28} & \textbf{92.15}
                                 & \textbf{93.49} & \textbf{80.11}
                                 & \textbf{70.13} & \textbf{55.42} \\
\Xhline{1.5pt}
\end{tabular}
\end{table}

Table~\ref{tab:ocsr-metric} reports the evaluation results of all systems on the latest revised version of MolRecBench-Wild and its three difficulty subsets. MinerU.Chem's molecular structure recognition module, implemented using GTR-VL-1.4.13, achieves a SMILES exact-match accuracy of 93.02\% on the SMILES-evaluable subset ($N=2{,}392$) and a graph accuracy of 79.66\% on all graph-annotated samples ($N=5{,}024$). These results outperform the best comparison systems evaluated in this report by 18.15 percentage points in SMILES exact-match accuracy (GPT-5.6-Sol, 74.87\%) and 42.1 percentage points in graph accuracy (Gemini-3.5-flash-thinking, 37.56\%). In the difficulty-stratified evaluation, MinerU.Chem achieves SMILES exact-match and graph accuracies of 98.28\% and 92.15\%, respectively, on subset A; 93.49\% and 80.11\% on subset B; and 70.13\% and 55.42\% on subset C.

Performance decreases from subset A to subset C as visual complexity increases and chemical-semantic challenges are introduced. Subset C remains the most challenging, particularly under graph accuracy, which requires exact agreement of molecular topology and the evaluated atom- and bond-level attributes.

In Table~\ref{tab:ocsr-metric}, Full denotes the complete evaluation set, while A, B, and C denote the three difficulty subsets described above. SMILES exact-match accuracy and graph accuracy are computed on samples with the corresponding annotations. All values are percentages. A dash indicates that the corresponding metric was not evaluated because the system does not support the required output format or no compatible prediction output was available.

Representative predictions and their visualizations are provided in the Support Information.

\section{Reaction Scheme Parsing}
\label{sec:6-reaction-scheme-parsing}

Reaction scheme parsing converts reaction diagrams in documents into machine-readable records containing reactants, products, and reaction conditions. These records are organized in the Reaction Summary List. The task follows the core setting of RxnScribe-style reaction diagram parsing~\citep{qian2023rxnscribe}, focusing on structured extraction rather than broader reaction understanding.

Reaction scheme parsing is integrated with document-level molecule organization. Whenever possible, reactants and products are linked to entries in the Molecule Summary List, allowing each reaction component to be traced to its page number, within-page bounding box, identifier, and recognized structure. These links preserve both the composition of each reaction and the corresponding source evidence in the original document.

The current module supports common single-step reaction schemes and a subset of multi-step schemes. Its scope is limited to extracting structured records from reaction diagrams; it does not aim to infer reaction mechanisms or recommend reaction conditions. The training data builds on the team's prior work, RxnCaption~\citep{song2026rxncaption}, and covers reactants, products, and conditions in typical reaction schemes, together with common layout variations.

\section{Applications and Future Directions}
\label{sec:7-applications-and-future-directions}

MinerU.Chem is positioned as a structured data production layer for organic chemistry documents. Its goal is to convert molecular structure depictions, reaction schemes, and related information in papers and patents into traceable and verifiable structured data. Based on its current capabilities, MinerU.Chem supports three primary application scenarios.

The first scenario is \textbf{synthetic route information extraction from natural product total synthesis and medicinal chemistry papers}. Such literature often revolves around a target molecule and contains long synthetic routes with multiple intermediates, reaction steps, and conditions. MinerU.Chem can extract molecular structures, reactants, products, and selected reaction conditions from route diagrams, providing structured inputs for reconstructing complete multi-step synthetic pathways.

The second scenario is \textbf{reaction and mechanism diagram extraction from organic methodology papers}. Methodology papers often contain reaction schemes, mechanism diagrams, condition screening tables, and substrate scope tables. MinerU.Chem can currently recognize molecular structure depictions and reaction schemes and organize molecules and intermediates appearing in mechanism diagrams. However, interpreting more complex mechanistic semantics---such as electron-pushing arrows and catalytic-cycle logic---and extracting reaction information from condition-screening and substrate-scope tables require further integration with table understanding and document-level semantic parsing. Understanding, parsing, and structuring complex chemical tables and mechanism diagrams is an important direction for future work.

The third scenario is \textbf{molecular property dataset construction}. For tasks such as molecular property prediction, virtual screening, and molecular design, MinerU.Chem can extract molecular structures and their associated textual context from papers, providing a basis for constructing molecule--property datasets. However, many papers describe chemical series using molecular templates, Markush structures, or R-group substitution tables. Processing such cases requires joint understanding of generic scaffolds, substituent definitions, table fields, and structure-instantiation relationships. The current version does not yet support fully automated expansion of Markush structures and R-group substitution patterns; this remains an important direction for future research and system development.

Overall, MinerU.Chem is intended to complement rather than replace expert chemical judgment. It provides an automated step for converting raw documents into candidate structured data, which experts can inspect, correct, and validate within a more efficient data-production workflow. We welcome collaboration on synthetic route reconstruction, reaction and mechanism diagram extraction, and molecular property dataset construction. MinerU.Chem is available as an online service at \href{https://mineru.net/OpenSourceTools/Extractor}{MinerU}. Interested teams are welcome to try the system and contact us.

\section*{Acknowledgments}

This project was supported by Shanghai Artificial Intelligence Laboratory.

\clearpage
\bibliographystyle{plainnat}
\bibliography{paper}

\clearpage

\section*{Support Information}
\label{sec:support-information}

\subsection*{A. Visualization of Molecule Recognition Results}

\begin{figure}[h]
    \centering
    \includegraphics[width=1\linewidth]{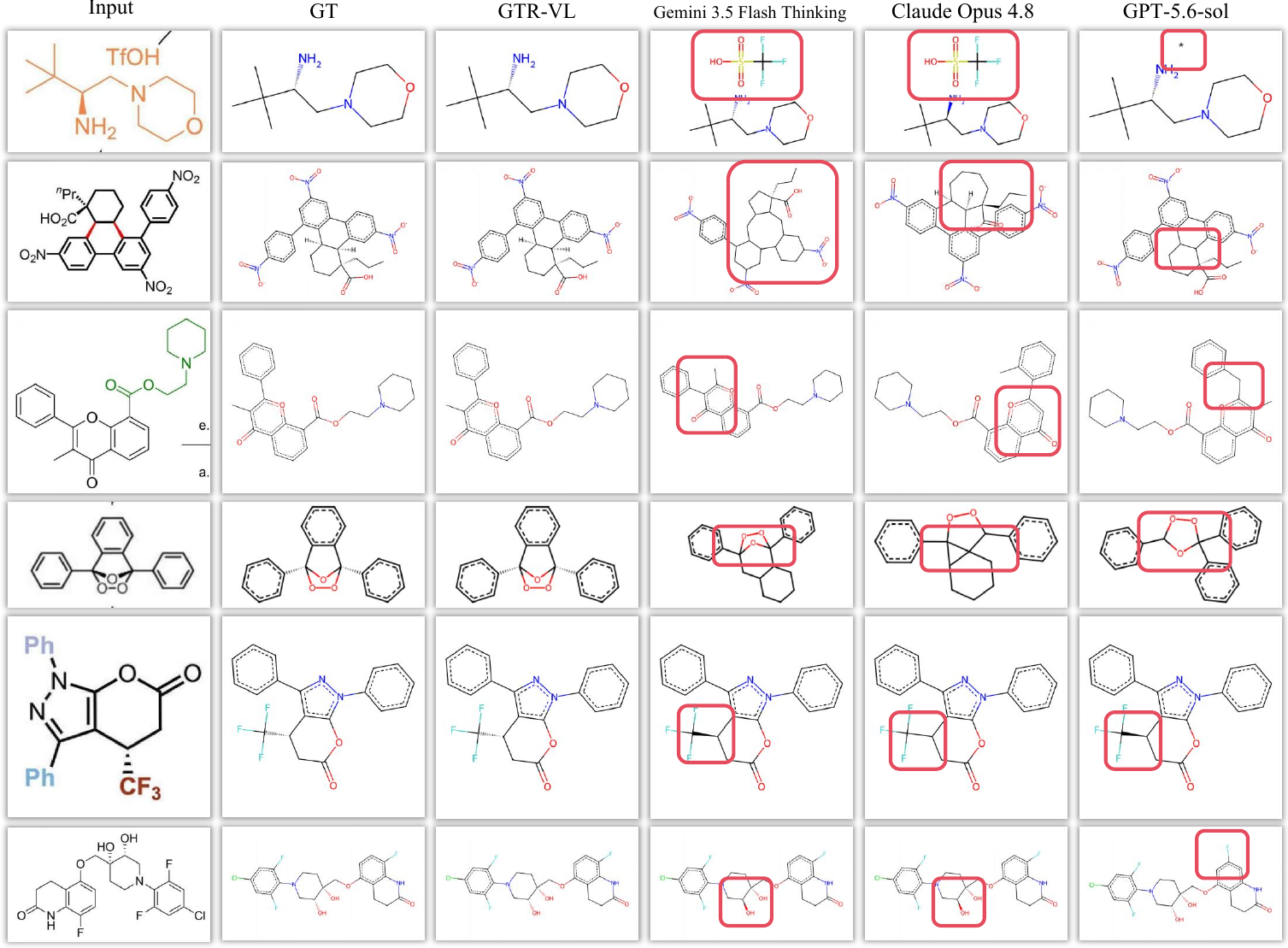}
    \caption{Comparison of SMILES predictions across the evaluated systems.}
    \label{fig:ocsr-vis-smiles-1}
\end{figure}

Figure~\ref{fig:ocsr-vis-smiles-1} presents representative SMILES predictions from the evaluated systems. 

\begin{figure}[h]
    \centering
    \includegraphics[width=1\linewidth]{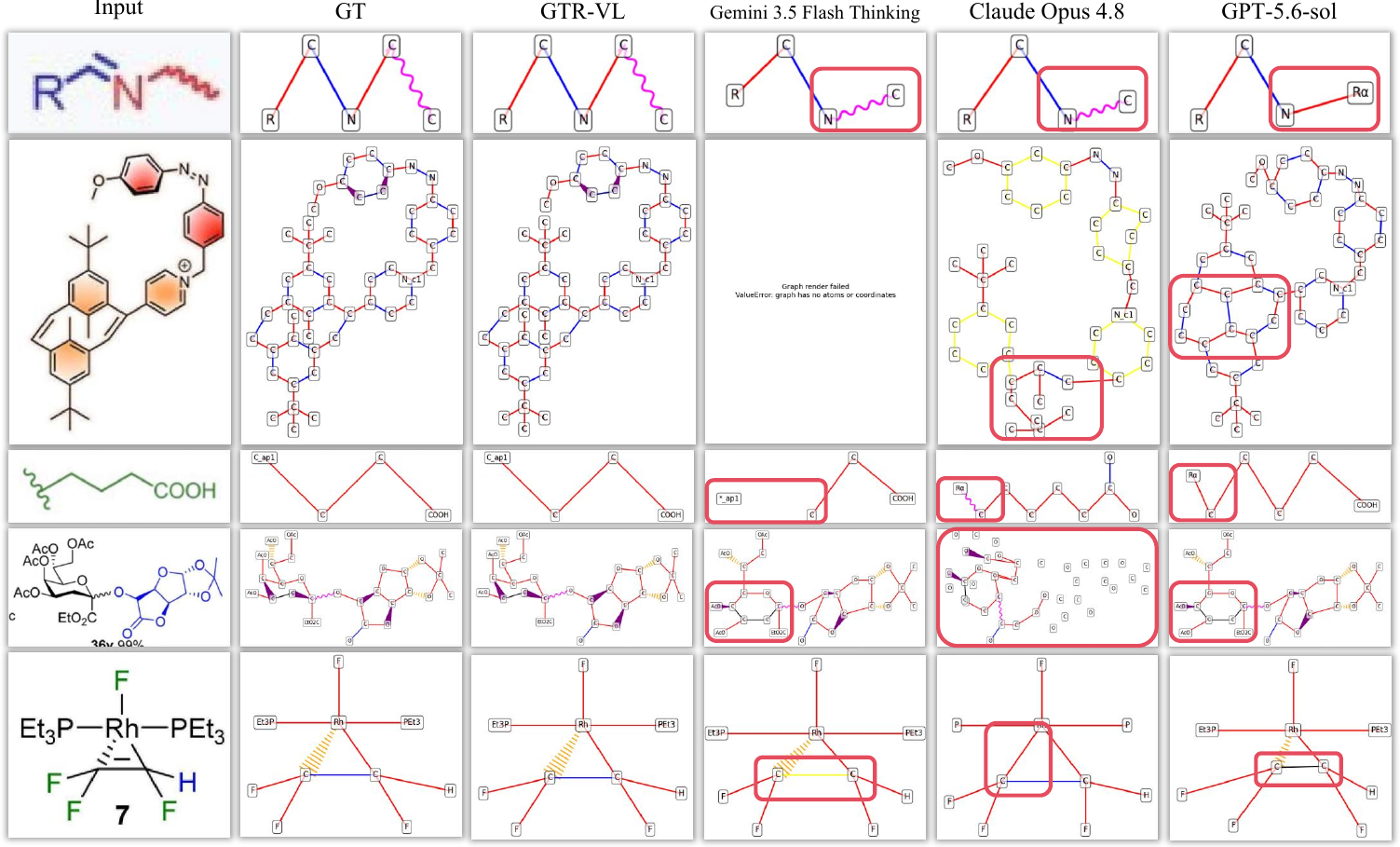}
    \caption{Comparison of molecular graph predictions across the evaluated systems.}
    \label{fig:ocsr-vis-graph-1}
\end{figure}

\begin{figure}[h]
    \centering
    \includegraphics[width=1\linewidth]{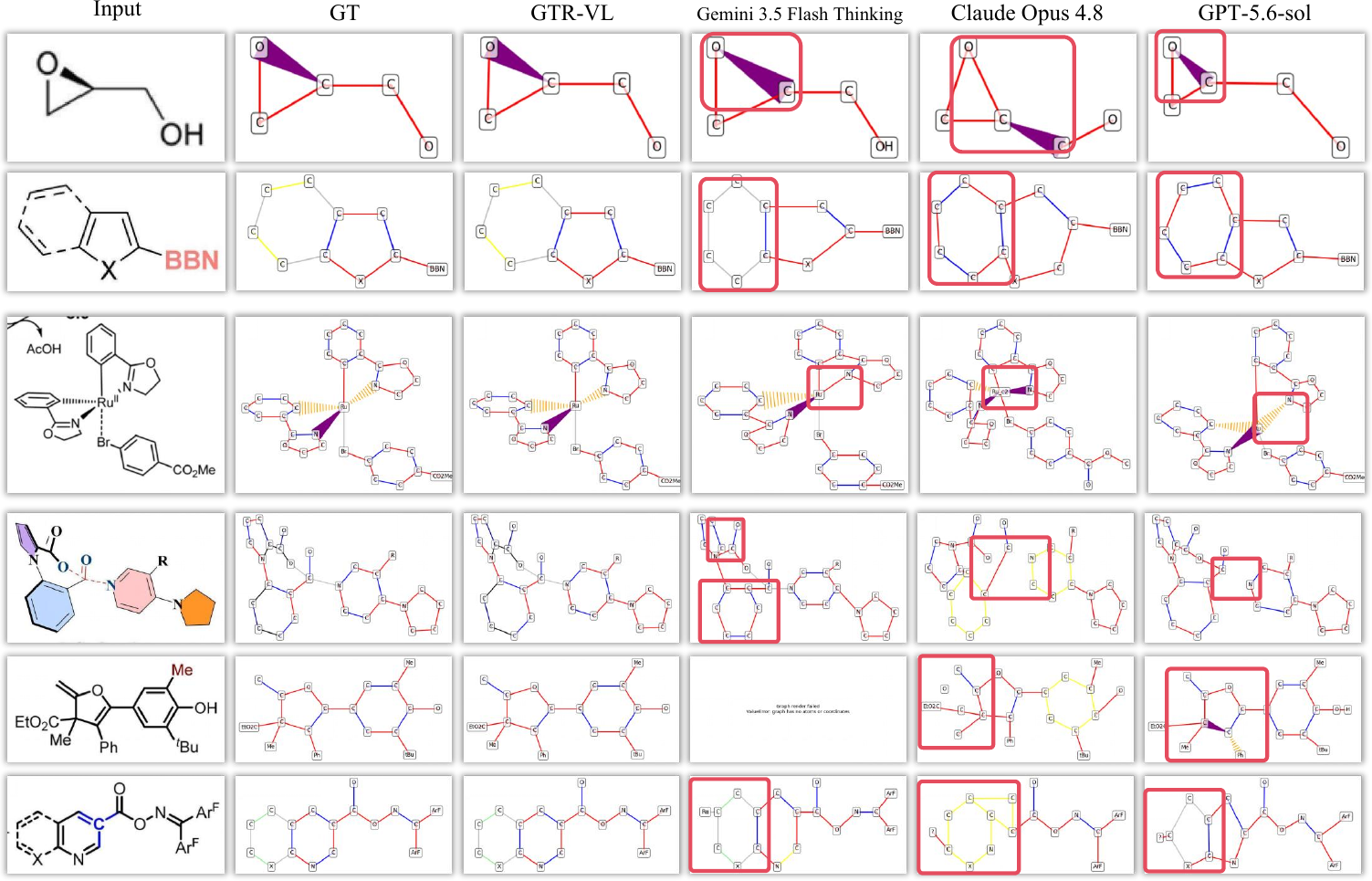}
    \caption{Comparison of molecular graph predictions across the evaluated systems.}
    \label{fig:ocsr-vis-graph-2}
\end{figure}

\begin{figure}[h]
    \centering
    \includegraphics[width=1\linewidth]{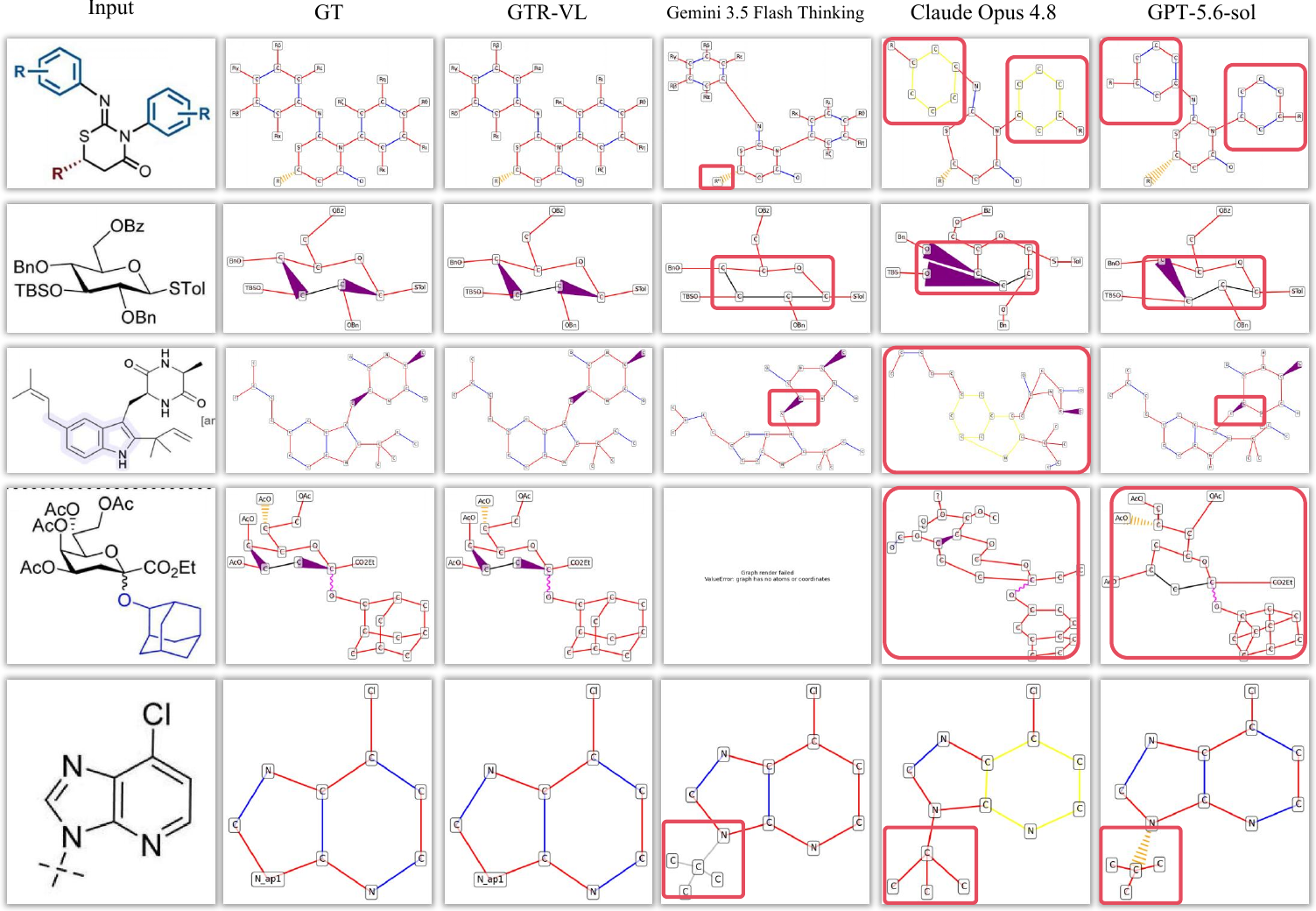}
    \caption{Comparison of molecular graph predictions across the evaluated systems.}
    \label{fig:ocsr-vis-graph-3}
\end{figure}

\begin{figure}[h]
    \centering
    \includegraphics[width=1\linewidth]{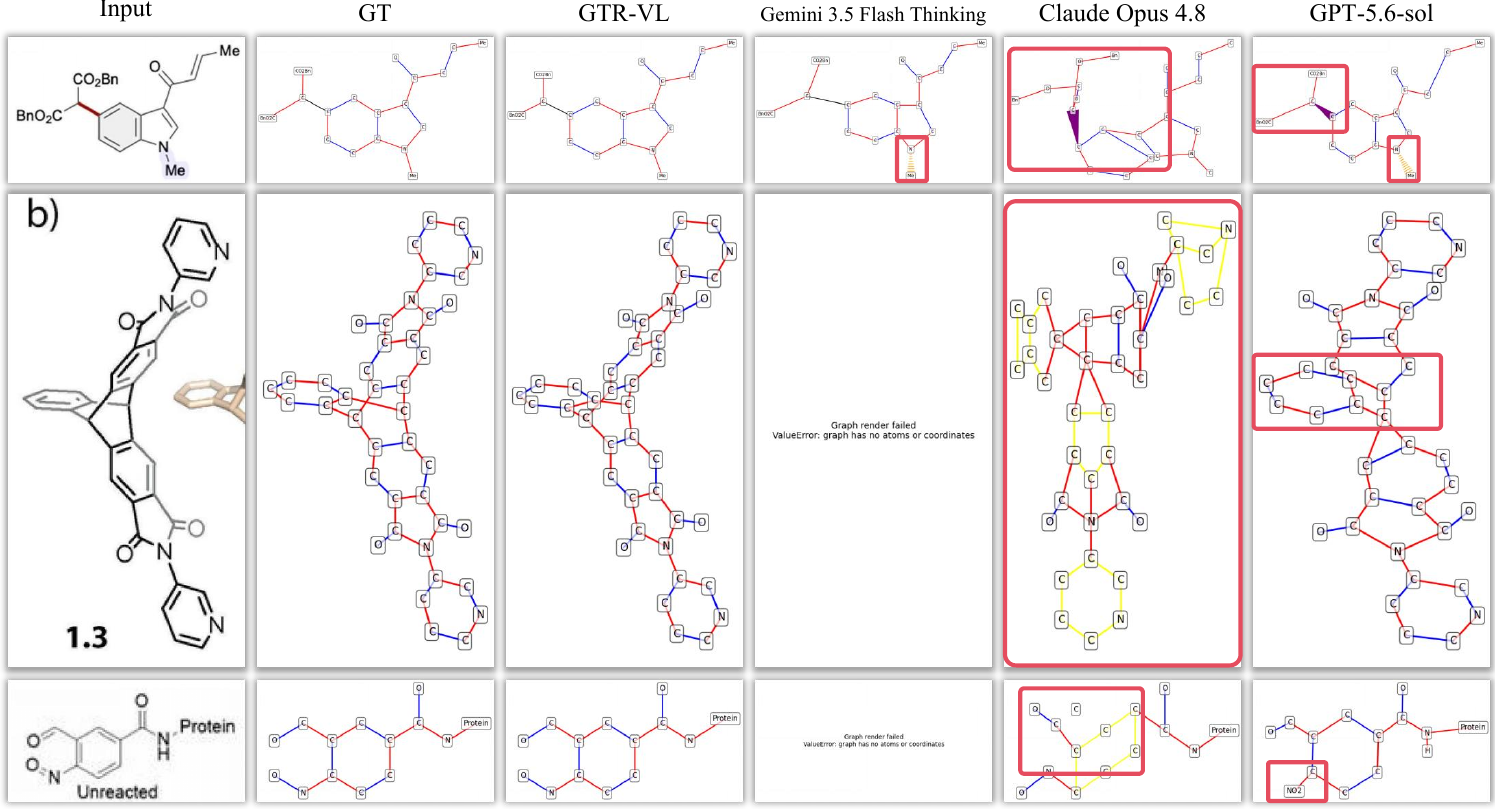}
    \caption{Comparison of molecular graph predictions across the evaluated systems.}
    \label{fig:ocsr-vis-graph-4}
\end{figure}

Figures~\ref{fig:ocsr-vis-graph-1}--\ref{fig:ocsr-vis-graph-4} present representative molecular graph predictions.
Table~\ref{tab:bond-color-mapping} lists all the bond types, examples, and corresponding visualization color appeared in the comparison of graph predictions.
Additional atom-level properties are displayed directly in the atom labels. Nonzero formal charges and valences are appended using the suffixes \texttt{\_c\{charge\}} and \texttt{\_v\{valence\}}, respectively.
Nonempty attachment points are appended using the suffix \texttt{\_ap\{attachment point\}}. These annotations are drawn as black text inside semi-transparent white rounded boxes. Radical information is retained in the graph representation but is not currently rendered as a separate visual annotation.

\begin{table}[!t]
\centering
\caption{All the bond types, colors, and examples appeared in the visualization results.}
\label{tab:bond-color-mapping}
\begin{tabular}{lcc|lcc}
\toprule
Type & Example & Color & Type & Example & Color \\
\midrule
Single bond                 & \raisebox{-0.5\height}{\includegraphics[width=1.5cm]{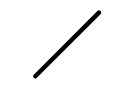}}                        & \textcolor{red}{Red}                      & 
Bold bond                   & \raisebox{-0.5\height}{\includegraphics[width=1.5cm]{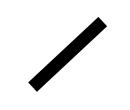}}                   & \textcolor{black}{Black}                  \\
Double bond                 & \raisebox{-0.5\height}{\includegraphics[width=1.5cm]{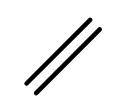}}                        & \textcolor{blue}{Blue}                    & 
Hashed bond                 & \raisebox{-0.5\height}{\includegraphics[width=1.5cm]{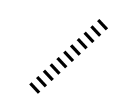}}                   & \textcolor{gray}{Gray}                    \\
Triple bond                 & \raisebox{-0.5\height}{\includegraphics[width=1.5cm]{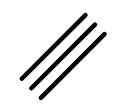}}                        & \textcolor{green}{Green}                  & 
Hollow wedge bond           & \raisebox{-0.5\height}{\includegraphics[width=1.5cm]{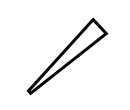}}                  & \textcolor{pink}{Pink}                    \\
Aromatic bond               & \raisebox{-0.5\height}{\includegraphics[width=1.5cm]{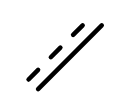}}                      & \textcolor{yellow!70!orange}{Yellow}      & 
Bold double bond            & \raisebox{-0.5\height}{\includegraphics[width=1.5cm]{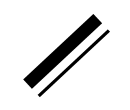}}                   & \textcolor{blue!70!black}{Dark blue}      \\
Solid wedge bond            & \raisebox{-0.5\height}{\includegraphics[width=1.5cm]{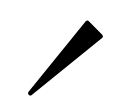}}                   & \textcolor{purple}{Purple}                & 
Dashed double bond          & \raisebox{-0.5\height}{\includegraphics[width=1.5cm]{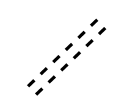}}                 & \textcolor{green!45!white}{Light green}   \\
Dashed wedge bond           & \raisebox{-0.5\height}{\includegraphics[width=1.5cm]{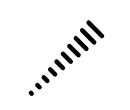}}                  & \textcolor{orange}{Orange}                & 
Dashed triple bond          & \raisebox{-0.5\height}{\includegraphics[width=1.5cm]{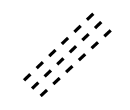}}                 & \textcolor{green!50!black}{Dark green}    \\
Any bond                    & \raisebox{-0.5\height}{\includegraphics[width=1.5cm]{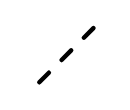}}                           & \textcolor{gray}{Silver gray}             & 
Dashed dative bond          & \raisebox{-0.5\height}{\includegraphics[width=1.5cm]{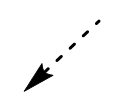}}                 & \textcolor{cyan!60!black}{Dark cyan}      \\
Dative bond                 & \raisebox{-0.5\height}{\includegraphics[width=1.5cm]{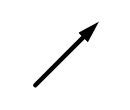}}                        & \textcolor{cyan}{Cyan}                    & 
Single-dashed double-solid  & \raisebox{-0.5\height}{\includegraphics[width=1.5cm]{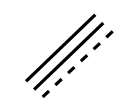}}    & \textcolor{olive}{Olive}                  \\
Hydrogen bond               & \raisebox{-0.5\height}{\includegraphics[width=1.5cm]{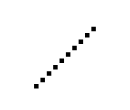}}                      & \textcolor{blue!45!cyan}{Light blue}      &      
Crossed double bond         & \raisebox{-0.5\height}{\includegraphics[width=1.5cm]{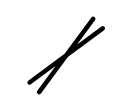}}                  & \textcolor{brown}{Brown}                  \\
Wavy bond                   & \raisebox{-0.5\height}{\includegraphics[width=1.5cm]{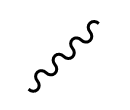}}                          & \textcolor{magenta}{Magenta}              & 
                            &                                                                               &                                           \\
\bottomrule
\end{tabular}
\end{table}


\end{document}